# 2
# Natural Science Paradigms

**Gerard Briscoe**
Digital Ecosystem Lab
Department of Media and Communications
London School of Economics and Political Science
http://personal.lse.ac.uk/briscoe/
g.briscoe@lse.ac.uk

**Suzanne Sadedin**
Department of Ecology and Evolutionary Biology
University of Tennessee
http://web.utk.edu/~ssadedin/
ssadedin@utk.edu

A primary motivation for research in Digital Ecosystems is the desire to exploit the self-organising properties of natural ecosystems. Ecosystems are thought to be robust, scalable architectures that can automatically solve complex, dynamic problems. However, the biological processes that contribute to these properties have not been made explicit in Digital Ecosystem research. Here, we introduce how biological properties contribute to the self-organising features of natural ecosystems. These properties include populations of evolving agents, a complex dynamic environment, and spatial distributions which generate local interactions. The potential for exploiting these properties in artificial systems is then considered.

## Introduction

Natural science is the study of the universe via rules or laws of natural order. The term natural science is also used to differentiate those fields using scientific method in the study of nature, in contrast with social sciences which use the scientific method applied to human behaviour. The fields of natural science are diverse, ranging from soil science to astronomy. We are by no means claiming that all these fields of study will provide paradigms for Digital Ecosystems. So, a brief summary of the relevant fields are shown below:

- ▶ Physics, the study of the fundamental constituents of the universe, the forces and interactions they exert on one another, and the results produced by these forces.
- ▶ Biology, the study of life
    - ▶ Ecology and Environmental science, the studies of the interrelationships of life and the environment.
- ▶ Chemistry, the study of the composition, chemical reactivity, structure, and properties of matter and with the (physical and chemical) transformations that they undergo.

The further one wishes to take the analogy of the word ecosystem in Digital Ecosystems, the more one has to consider the relevance of the fields of natural science, because our focus is in creating the digital counterpart of biological ecosystems.

# The Biology of Digital Ecosystems [7]

Is mimicking nature the future for information systems? A key challenge in modern computing is to develop systems that address complex, dynamic problems in a scalable and efficient way. One approach to this challenge is to develop Digital Ecosystems, artificial systems that aim to harness the dynamics that underlie the complex and diverse adaptations of living organisms. However, the connections between Digital Ecosystems and their biological counterparts have not been closely examined. Here, we consider how properties of natural ecosystems influence functions that are relevant to developing Digital Ecosystems to solve practical problems. This leads us to suggest ways in which concepts from ecology can be used to combine biologically inspired techniques to create an applied Digital Ecosystem.

The increasing complexity of software [17] makes designing and maintaining efficient and flexible systems a growing challenge. Many authors argue that software development has hit a complexity wall which can only be overcome by automating the search for new algorithms [23]. Natural ecosystems possess several properties that may be useful in such automated systems. These properties include self-organisation, self-management, scalability, the ability to provide complex solutions, and automated composition of these complex solutions.

A natural ecosystem consists of a community of interacting organisms in their physical environment. Several fundamental properties influence the structure and function of natural ecosystems. These include agent population dynamics, spatial interactions, evolution, and complex, changing environments. We examine these properties, and the role they may play in an applied Digital Ecosystem. We suggest that several key features of natural ecosystems have not been fully explored in existing Digital Ecosystems, and discuss how mimicking these feature may assist in developing robust, scalable self-organising architectures.

Arguably the most fundamental differences between biological and Digital Ecosystems lie in the motivation and approach of researchers. Biological ecosystems are ubiquitous natural phenomena, whose maintenance is crucial to our survival. The performance of natural ecosystems is often measured in terms of their stability, complexity and diversity. In contrast, Digital Ecosystems as defined here are technology engineered to serve specific human purposes. The performance of a Digital Ecosystem, then, can be judged relative to the function it is designed to perform. In many cases, the purpose of a Digital Ecosystem is to solve dynamic problems in parallel with high efficiency. This criterion may be related only indirectly to complexity, diversity and stability, an issue we shall examine further.

Genetic algorithms are the subset of evolutionary computation that uses natural selection to evolve solutions. Starting with a set of arbitrarily chosen possible solutions, selection, replication, recombination and mutation are applied iteratively. Selection is based on conformation to a fitness function which is determined by the specific problem of interest. Over time, better solutions to the problem can thus evolve. Because they are intended to solve problems by evolving solutions, applied Digital Ecosystems are likely to incorporate some form of genetic algorithm. However, we suggest that a Digital Ecosystem should also incorporate additional features which give it a closer resemblance to natural ecosystems. Such features might include complex, dynamic fitness functions, a distributed or network environment, and self-organisation arising from interactions among agents and their environment. These properties are discussed further below.

## Fitness Landscapes and Agents

As described above, an ecosystem comprises both an environment and a set of interacting, reproducing entities (or agents) in that environment. In biological ecosystems, the environment acts as a set of physical and chemical constraints on reproduction and survival. These constraints can be considered in abstract using the metaphor of the fitness landscape [26]. Individuals are represented in the fitness landscape as solutions to the problem of survival and reproduction.

All possible solutions are distributed in a space whose dimensions are the possible properties of individuals. An additional dimension, height, indicates the relative fitness (in terms of survival and reproduction) of each solution. The fitness landscape is envisaged as a rugged, multidimensional landscape of hills, mountains and valleys, because individuals with certain sets of properties are fitter than others.

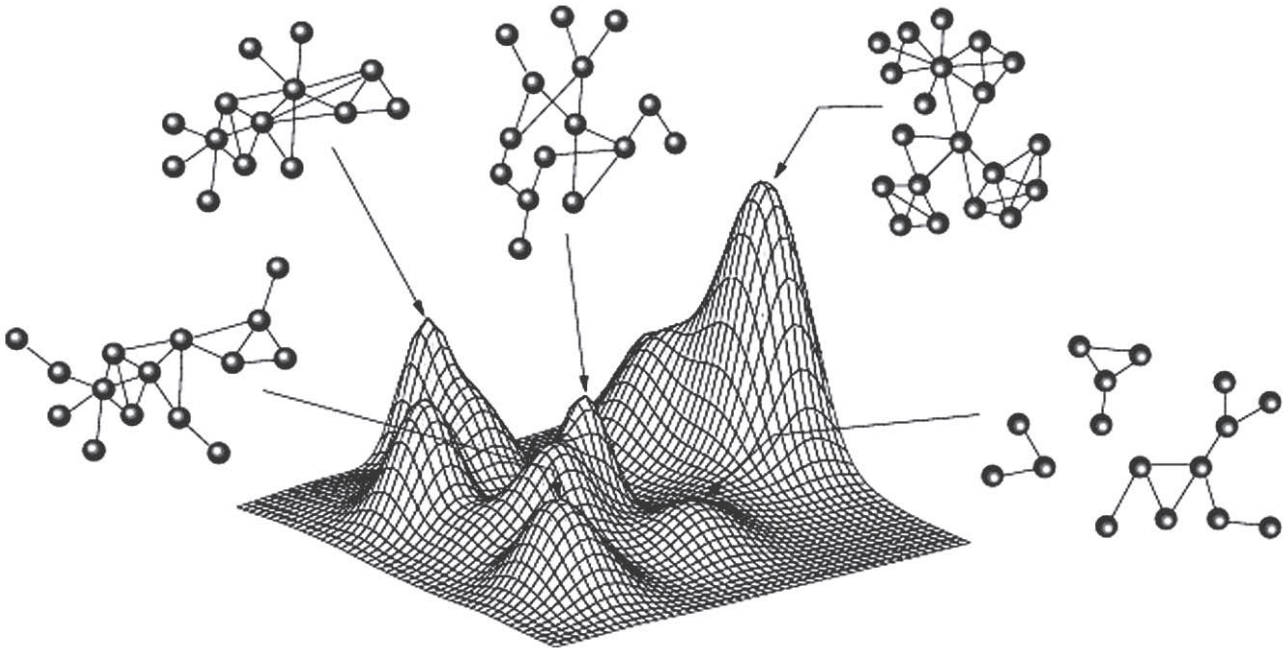

**Fig. 1**

A hypothetical software fitness landscape. Peaks correspond to sub-optimal solutions. We can represent software development as a walk through the landscape towards these peaks. The roughness of the landscape indicates how difficult is to reach an optimal software design. [24]

In biological ecosystems, fitness landscapes are virtually impossible to identify. This is both because there are large numbers of possible traits that can influence individual fitness, and because the environment changes over time and space. In contrast, within a digital environment, it is normally possible to specify explicitly the constraints that act on individuals in order to evolve solutions that perform better within these constraints.

Within genetic algorithms, exact specification of a fitness landscape or function is common practice. However, within a Digital Ecosystem, the ideal constraints are those that allow solution populations to evolve to meet user needs with maximum efficiency. User needs will change from place to place and time to time. In this sense the Digital Ecosystem fitness landscape is complex and dynamic, and more like that of a biological ecosystem than like that of a traditional genetic algorithm. The designer of a Digital Ecosystem therefore faces a double challenge: firstly, to specify rules that govern the shape of the fitness landscape in a way that meaningfully maps landscape dynamics to user requests, and secondly, to evolve solution populations within this space that are diverse enough to solve disparate problems, complex enough to meet user needs, and efficient enough to be superior to those generated by other means.

The agents within a Digital Ecosystem are like biological individuals in the sense that they reproduce, vary, interact, move and die. Each of these properties contributes to the dynamics of the ecosystem. However, the way in which these individual properties are encoded may vary substantially depending on the purpose of the system.

## Networks and Spatial Dynamics

A key factor in the maintenance of diversity in natural ecosystems is spatial interactions. Several modelling systems have been used to represent spatial interactions. These include metapopulations, diffusion models, cellular automata and agent-based models (termed individual-based models in ecology). The broad predictions of these diverse models are in good agreement. At local scales, spatial interactions favor relatively abundant species disproportionately. However, at a wider scale, this effect can preserve diversity, because different species will be locally abundant in different places. The result is that even in homogeneous environments, population distributions tend to form discrete, long-lasting patches that can resist invasion by superior competitors [11]. Population distributions can also be influenced by environmental variations such as barriers, gradients and patches. The possible behavior of spatially distributed ecosystems is so diverse that scenario-specific modelling is necessary to understand any real system [9]. Nonetheless, certain robust patterns are observed. These include the relative abundance of species (which consistently follows a roughly log-normal relationship) [3] and the relationship between geographic area and the number of species present (which follows a power law) [1]. The reasons for these patterns are disputed: both spatial extensions of simple Lotka-Volterra competition models [14], and more complex ecosystem models [22], are capable of generating them.

Landscape connectivity plays an important part in ecosystems. When the density of habitat within an environment falls below a critical threshold, widespread species may fragment into isolated populations. Fragmentation can have several consequences. Within populations, these effects include loss of genetic diversity and harmful inbreeding [10]. At a broader scale, isolated populations may diverge genetically, leading to speciation. From an information theory perspective, this phase change in landscape connectivity can mediate global and local search strategies [13]. In a well-connected landscape, selection favors the globally superior, and pursuit of different evolutionary paths is discouraged, potentially leading to premature convergence. When the landscape is fragmented, populations may diverge, solving the same problems in different ways. Recently, it has been suggested that the evolution of complexity in nature involves repeated landscape phase changes allowing selection to alternate between local and global search [12].

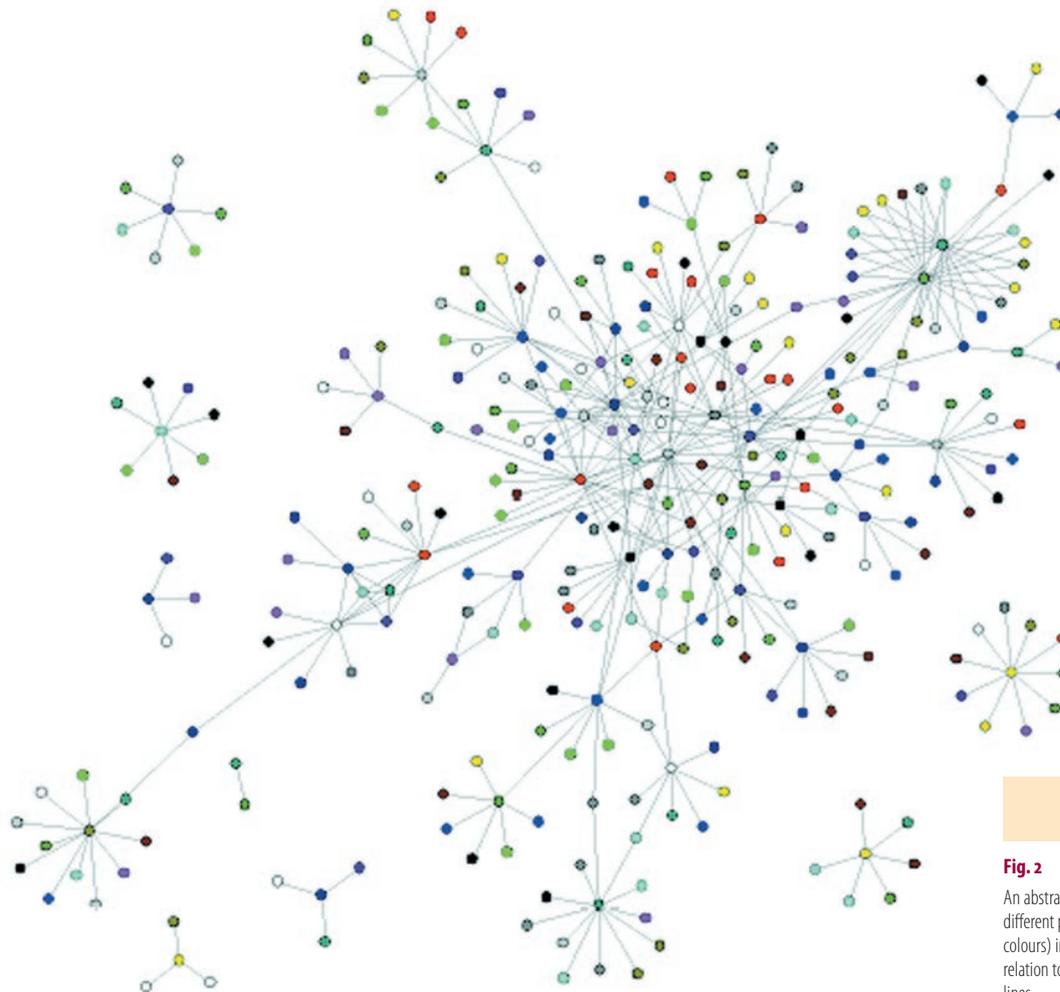

**Fig. 2**

An abstract view of an ecosystem, showing different populations (by the different colours) in different spatial areas, and their relation to one another by the connecting lines.

In a digital context, we can simulate spatial interactions by creating a distributed system that consists of a set of interconnected locations. Agents can migrate between connected locations with low probability. Depending on the how the connections between locations were organised, such Digital Ecosystems might have dynamics closely parallel to spatially explicit models, diffusion models, or metapopulations [9]. In all of these systems, the spatial dynamics are relatively simple compared with those seen in real ecosystems, which incorporate barriers, gradients and patchy environments at multiple scales in continuous space. Another alternative in a Digital Ecosystem is to apply a spatial system that has no geometric equivalent. These could include, for example, small world networks and geographic systems that evolve by Hebbian learninonig. We will discuss an example of a non-geometric spatial network, and some possible reasons for using this approach, in a later section.

## Selection and Self-Organisation

The major hypothetical advantage of Digital Ecosystems over other complex organisational models is their potential for dynamic adaptive self-organisation. However, for the solutions evolving in Digital Ecosystems to be useful, they must not only be efficient in a computational sense, they must also solve meaningful problems. That is, the fitness of agents must translate in some sense to real-world usefulness as demanded by users.

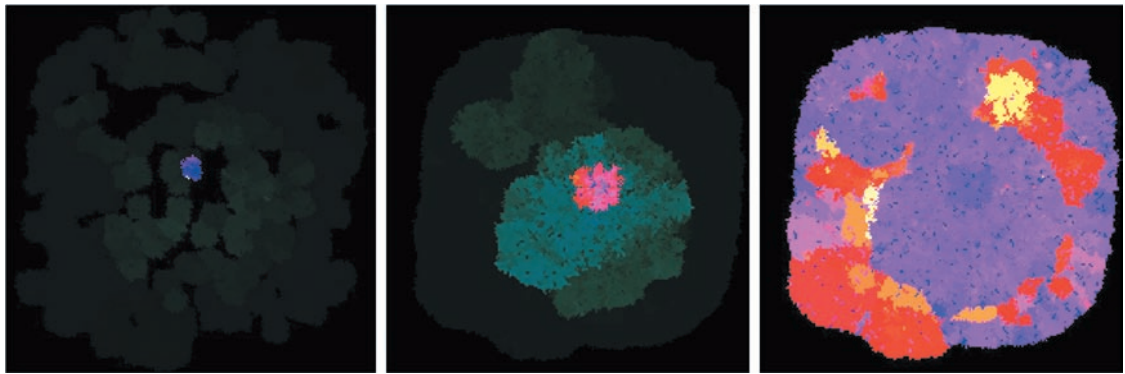

**Fig. 3**
An evolving population of digital organisms in a virtual petri dish at three time steps, showing the self-organistion of the population undergoing selection. The colour shows the genetic variability of the digital organisms, with the genetically identical being the same colour. [20]

Constructing a useful Digital Ecosystem therefore requires a balance between freedom of the system to self-organise, and constraint of the system to generate useful solutions. These factors must be balanced because the more the system's behavior is dictated by internal dynamics of the system, the less it may respond to fitness criteria imposed by users. At one extreme, when system dynamics are mainly internal, agents may evolve that are good at survival and reproduction within the digital environment, but useless in the real world. At the other extreme, where the users' fitness criteria overwhelmingly dictate function, we suggest that dynamic exploration of solution space and complexity are likely to be limited.

The reasoning behind this argument is as follows. Consider a multidimensional solution space which maps to a rugged fitness landscape. In this landscape, competing solution lineages will gradually become extinct through chance processes. Consequently, the solution space explored becomes smaller over time as the population adapts and diversity of solutions decreases. Ultimately, all solutions may be confined to a small region of solution space. In a static fitness landscape, this situation is not undesirable because the surviving solution lineages will usually be clustered around an optimum. However, if the fitness landscape is dynamic, the location of optima varies over time. Should lineages become confined to a small area of solution space, subsequent selection will locate only optima that are near this area. This is undesirable if new, higher optima arise that are far from pre-existing ones. A related issue is that complex solutions are less likely to be found by chance than simple ones. Complex solutions can be visualised as sharp, isolated peaks on the fitness landscape. Especially in the case of a dynamic landscape, these peaks are most likely to be found when the system explores solution space widely. Consequently, a self-organising mechanism other than the fitness criteria of users is required to maintain diversity among competing solutions in the Digital Ecosystem.

## Stability and Diversity in Complex Adaptive Systems

Ecosystems are often described as complex adaptive systems (CAS). That is, they are systems comprised of diverse, locally interacting components that are subject to selection. Other complex adaptive systems include brains, individuals, economies, and the biosphere. All are characterised by hierarchical organisation, continual adaptation and novelty, and non-equilibrium dynamics. These properties lead to behavior that is non-linear, historically contingent, subject to thresholds, and contains multiple basins of attraction [15].

In the above sections we have advocated Digital Ecosystems that include agent populations evolving by natural selection in distributed environments. Like real ecosystems, digital systems designed in this way fit the definition of complex adaptive systems. The features of these systems, especially non-linearity and non-equilibrium dynamics, offer both advantages and hazards for adaptive problem-solving. The major hazard is that the dynamics of complex adaptive systems are intrinsically hard to predict due to self-organisation. This observation implies that designing a useful Digital Ecosystem will be partly a matter of trial and error. The occurrence of multiple basins of attraction in complex adaptive systems suggests that even a system that functions well for a long period may at some point suddenly transition to a less desirable state. For example, in some types of system self-organising mass extinctions might result from interactions among populations, leading to temporary unavailability of diverse solutions. This concern may be addressed by incorporating negative feedback mechanisms at the global scale.

The challenges in designing an effective Digital Ecosystem are mirrored by the system's potential strengths. Nonlinear behavior provides the opportunity for scalable organisation and the evolution of complex hierarchical solutions. Rapid state transitions potentially allow the system to adapt to sudden environmental changes with minimal loss of functionality.

A key question for designers of Digital Ecosystems is how the stability and diversity properties of natural ecosystems map to performance measures in digital systems. For a Digital Ecosystem, the ultimate performance measure is user satisfaction, a system-specific property. However, assuming that the motivation for engineering a Digital Ecosystem is the development of scalable, adaptive solutions to complex dynamic problems, certain generalisations can be made. Sustained diversity, as discussed above, is a key requirement for dynamic adaptation. In the Digital Ecosystem, diversity must be balanced against adaptive efficiency because maintaining large numbers of poorly-adapted solutions is costly. The exact form of this tradeoff will be guided by the specific requirements of the system in question. Stability, as discussed above, is likewise a tradeoff: we want the system to respond to environmental change with rapid adaptation, but not to be so responsive that mass extinctions deplete diversity or sudden state changes prevent control.

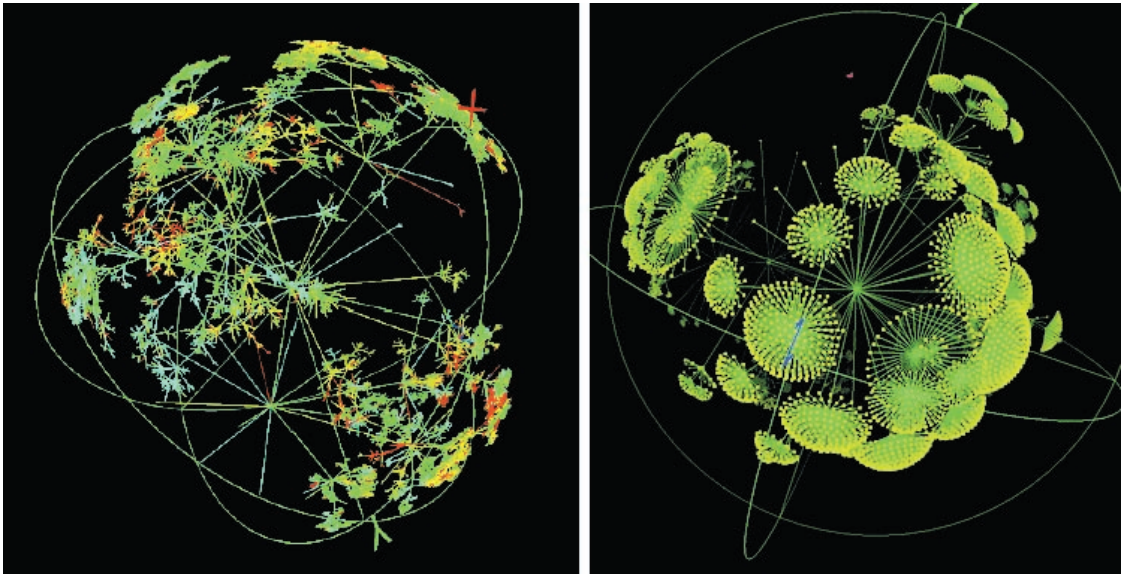

**Fig. 4**

Left: An abstract view of an ecosystem, showing the diversity of different populations by the different colours.

Right: An abstract view of population diversity within a single habitat, with space between points showing genetic diversity.

## Evolutionary Environment (EvE) Digital Ecosystem

The Digital Business Ecosystem (DBE) [21] is a proposed methodology for economic and technological innovation. Specifically, the DBE has a Digital Ecosystem software infrastructure for supporting a large number of interacting business users and services, called the EvE [4, 6, 5]. The individuals of a EvE are software agents that represent business services. These agents interact, evolve and adapt to a dynamic digital environment, thereby serving the ever-changing business requirements imposed by the economy.

The EvE is a two-level optimisation scheme inspired by natural ecosystems. A decentralised peer-to-peer network forms an underlying tier of distributed services. These services feed a second optimisation level based on genetic algorithms that operates locally on single peers (habitats) and is aimed at finding solutions satisfying locally relevant constraints. Through this twofold process, the local search is sped up and yields better local optima as the distributed optimisation provides prior sampling of the search space by making use of computations already performed in other peers with similar constraints.

The EvE is a Digital Ecosystem in which autonomous mobile agents represent various services (or compositions of services) offered by participating businesses. The abiotic environment is represented by a network of interconnected habitats nodes. Each connected enterprise has a dedicated habitat. Enterprises search for, and deploy, services in local habitats. Continuous and varying user requests for services provide a dynamic evolutionary pressure on the agents, which have to evolve to better satisfy those requests.

In natural ecosystems geography defines the connectivity between habitats. Nodes in the EvE do not have a default geographical topology to define connectivity. A re-configurable network topology is used instead, which is dynamically adapted on the basis of observed migration paths of the individuals within the EvE habitat network. Following the idea of Hebbian learning, the habitats which succesfully exchange individuals more often obtain stronger connections and habitats which do not succesfully exchange individuals will become less strongly connected. This way, a network topology is discovered with time, which reflects the structure of the business-sectors within the user base. The resulting

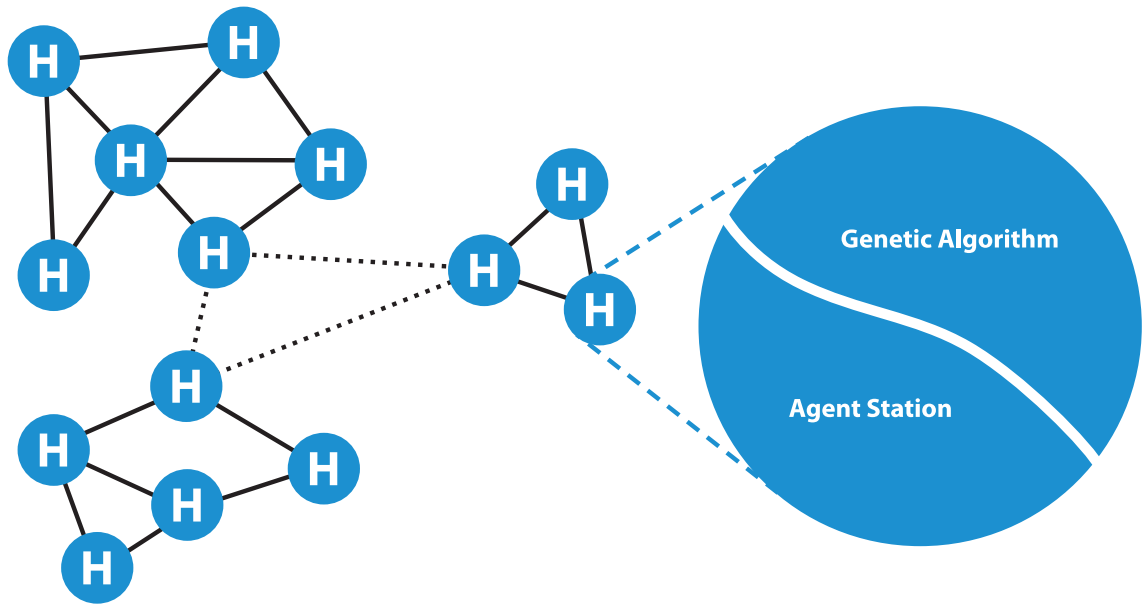

**Fig. 5**

Optimisation architecture: agents travel along the peer-to-peer connections that have the topology of a small-world network; in each node local optimisation is performed through a genetic algorithm, where the search space is defined by the agents contained in the node.

network will resemble the connectivity of the businesses within the user base, typically a small-world network in the case of Small & Medium Enterprises (SMEs) [27, 8, 19]. Such a network has many strongly connected clusters called sub-networks (quasi complete graphs) and a few connections between these clusters. Graphs with this topology have a very high clustering coefficient and small characteristic path lengths [25, 18], as shown in Figure 5.

In simulation we compared some of the EvE's dynamics to those of natural ecosystems, and the experimental results indicated that under simulation conditions the EvE behaves in some ways like a natural ecosystem. This suggests that incorporating ideas from theoretical ecology can contribute to useful self-organising properties in Digital Ecosystems. [7]

## Conclusions

By comparing and contrasting theoretical ecology with the anticipated requirements of Digital Ecosystems, we have examined how ecological features may emerge in some systems designed for adaptive problem solving. Specifically, we suggested that a Digital Ecosystem, like a real ecosystem, will usually consist of self-replicating agents that interact both with one another and with an external environment. Agent population dynamics and evolution, spatial and network interactions, and complex dynamic fitness landscapes will all influence the behaviour of these systems. Many of these properties can be understood via well-known ecological models [16, 14].

A further body of theory treats ecosystems as complex adaptive systems [15]. These models provide a theoretical basis for the occurrence of self-organisation in both digital and real ecosystems. Self-organisation results when interactions among agents and their environment giving rise to complex non-linear behavior. It is this property that provides the underlying potential for scalable problem-solving in a digital environment.